\documentclass{article} 
\usepackage{iclr2027_conference,times}
\usepackage{dsfont}
\usepackage{caption}
\usepackage{fancyhdr}

\usepackage{amsmath,amsfonts,bm}

\def\eqref#1{equation~\ref{#1}}

\def\1{\bm{1}}

\DeclareMathAlphabet{\mathsfit}{\encodingdefault}{\sfdefault}{m}{sl}
\SetMathAlphabet{\mathsfit}{bold}{\encodingdefault}{\sfdefault}{bx}{n}

\usepackage{multirow}
\usepackage{booktabs}
\usepackage[table]{xcolor}
\usepackage{hyperref}
\usepackage{url}
\usepackage{graphicx}
\usepackage{subcaption}

\title{T-LoopFormer: \textbf{T}oken-Level Elastic-Depth \textbf{Looped Transformers} with Dynamic Routing}

\author{
Mingqian Yu$^{1}$, \quad Wenpeng Zhang, \quad Shaobo Cui$^{2}$, \quad Peilin Zhao$^{2,}$ \thanks{: Corresponding author.} \\
$^{1}$Institute of Automation, Chinese Academy of Sciences, Beijing, China \\
$^{2}$School of Artificial Intelligence, Shanghai Jiao Tong University, Shanghai, China \\
\texttt{yumingqian2026@ia.ac.cn}, \texttt{zhangwenpeng0@gmail.com}, \texttt{shaobo.cui@sjtu.edu.cn}\\
\texttt{peilinzhao@sjtu.edu.cn}
}

\iclrfinalcopy 
\begin{document}
\maketitle
\fancyhf{} 
\renewcommand{\headrulewidth}{0pt} 

\begin{abstract}
Looped Transformers have recently demonstrated strong performance in both reasoning and language tasks by reusing a shared set of parameters across multiple iterations, achieving parameter efficiency without sacrificing representational power. Besides, looped Transformers perform inference directly in the latent space (latent reasoning) to reduce the number of tokens consumed during inference, thereby achieving improved sample efficiency. However, these models typically apply a fixed recursion depth uniformly to every token, leading to suboptimal compute allocation and leaving significant efficiency gains on the table. In this work, we propose \textbf{t}oken-level elastic-depth \textbf{looped transformers}: T-LoopFormer. T-LoopFormer uses \textbf{dynamic token-choice routing}, which enables each token to adaptively determine its own number of loop iterations based on its hidden state, which could improve the token generation accuracy. Moreover, we further introduce recursion-wise KV cache, which maintains an independent key-value cache for each recursion loop of active tokens; this design ensures that tokens at different depths only attend to their corresponding cached states, effectively enabling faster autoregressive decoding. Extensive experiments show that T-LoopFormer achieves robust performance on a wide range of language modeling and zero-shot reasoning tasks, and our model could reach the lowest decoding latency, which validates the effectiveness of token-choice router and recursion-wise KV cache.
Code is available at \url{https://github.com/YuMingQian1234/T-LoopFormer} 
\end{abstract}

\section{Introduction}
Transformers with parameter sharing, often called looped or recurrent Transformers, have emerged as an efficient and capable alternative to deep non–shared stacks across vision and natural language \citep{jeddi2026loopformer, dehghani2018universal, lan2019albert, geiping2026scaling, yu2026latent}. Notably, looped Transformers exhibit strong performance on a broad range of algorithmic and reasoning tasks when applied to language modeling \citep{geiping2026scaling, jeddi2026loopformer, saunshi2024inductive}. These models perform latent reasoning during inference, which allows elastic scaling of inference-time computational depth without increasing the parameter count, thereby improving computation efficiency and yielding better results on reasoning benchmarks and other downstream tasks and more computation in vanilla Transformers requires more layers, raising overhead \citep{geiping2026scaling, saunshi2024inductive}. However, a fixed number of loops is adopted by current approaches during training and inference. 

Although there are methods that can achieve elastics-depth looped Transformers \citep{geiping2026scaling, jeddi2026loopformer}, they do not provide \textbf{token‑level elastic-depth} control over the number of loops. When these models perform reasoning recursively in latent space, they apply a uniform number of unrolling loops to all tokens, regardless of their individual hidden state \citep{jeddi2026loopformer, geiping2026scaling, prairie2026parcae, xu2024expressive}. This monolithic strategy lacks token‑level elasticity, meaning that different tokens recurse the same loop counts. As a result, the accuracy of the generated token will decrease, and the model incurs significant and often unnecessary memory overhead during training, which is particularly problematic when scaling to longer sequences or resource‑constrained settings. Besides, because these models conduct recurrent reasoning in the latent space in decoding, they suffer from reduced inference throughput and consequently higher response latency \citep{wu2025parallel, dehghani2018universal}. To overcome these questions, we propose \textbf{t}oken-level elastic-depth \textbf{looped transformers}: \textbf{T-LoopFormer}.
\begin{figure}[htbp]
    \centering
    \begin{subfigure}[b]{0.50\textwidth}
        \centering
        \includegraphics[width=\textwidth, height=11cm, keepaspectratio]
            {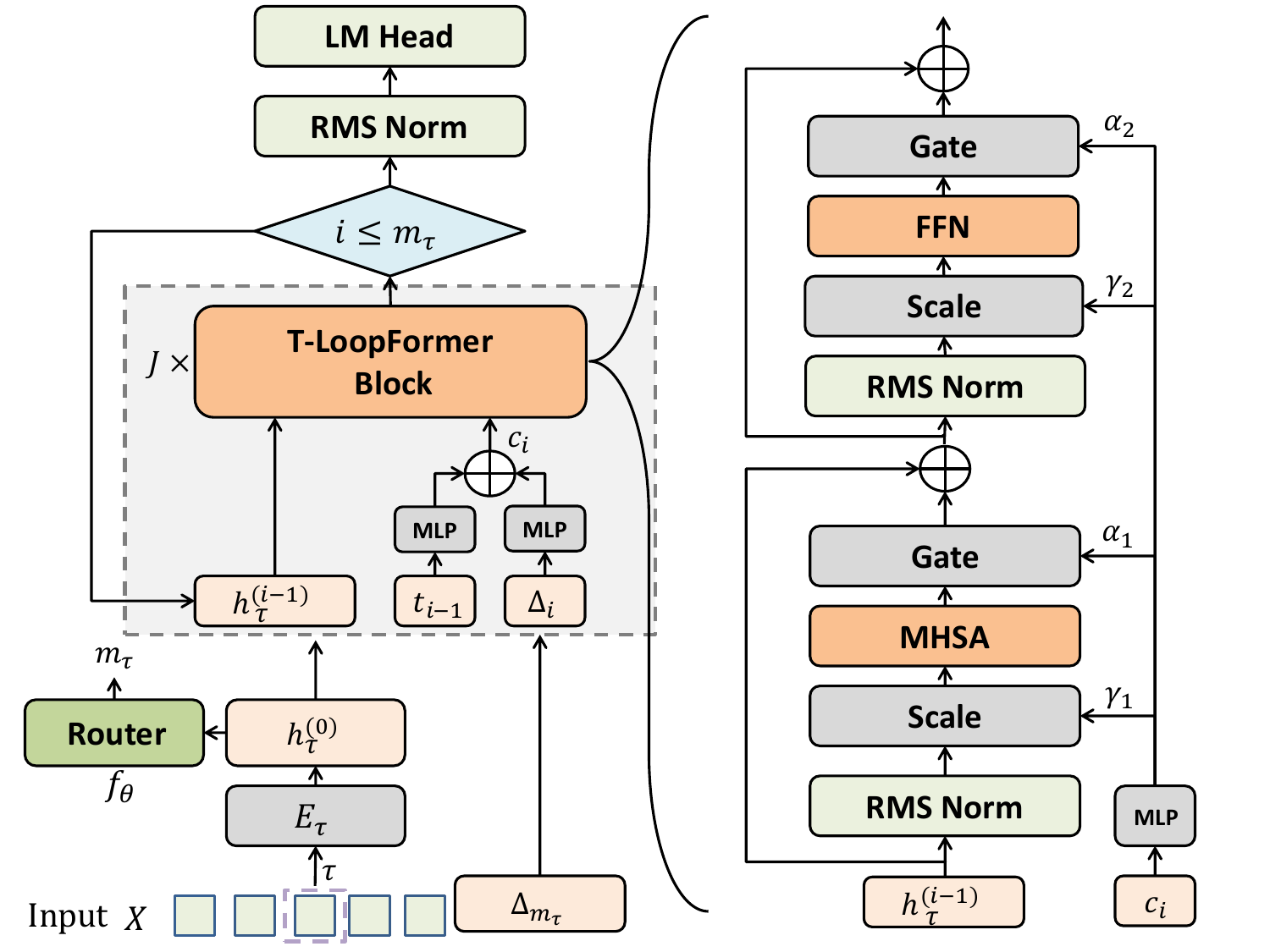}
        \caption{T-LoopFormer can achieve token-level
        elastic-depth through a dynamic token-choice router. The token-choice router determines the token's recurrent loops based on its hidden state, which can improve the token generation accuracy.}
        \label{fig:t_loopformer}
    \end{subfigure}
    \hfill
    \begin{subfigure}[b]{0.49\textwidth}
        \centering
        \includegraphics[width=\textwidth, height=11cm, keepaspectratio]
            {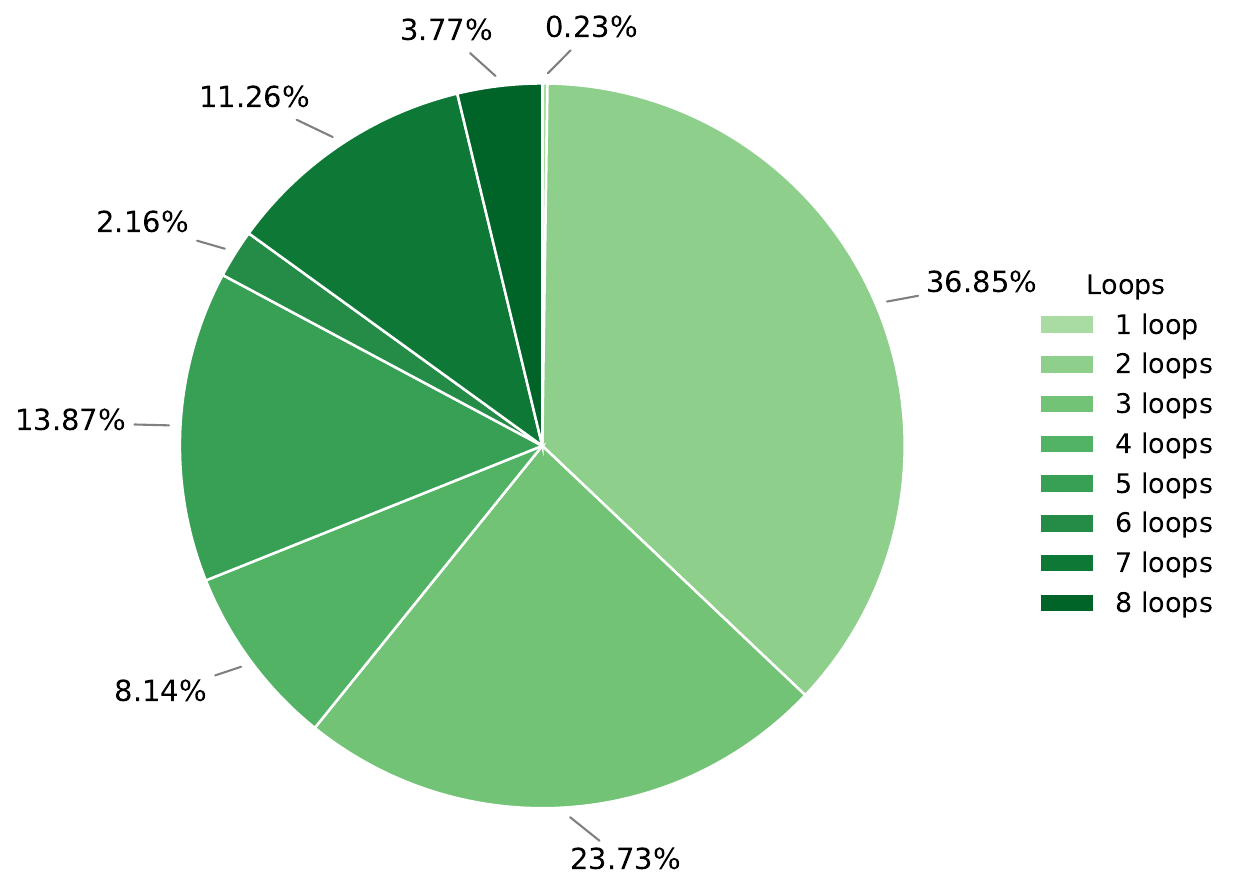}
        \caption{The recurrent depth ratio of T-LoopFormer on FineWeb-Edu validation set. The average recurrent depth is 3.69 and exhibits long-tailed distribution, which validates that different tokens require different recurrent loops. }
        \label{fig:loop_ratio_pie}
    \end{subfigure}
    \caption{(a) The architecture of T-LoopFormer. (b) The loop ratio of T-LoopFormer for different tokens on the FineWeb-Edu validation set at 24x FLOPs. }
    \label{model_and_loop_ratio}
\end{figure}
T-LoopFormer can achieve token-level elastic-depth through a dynamic token-choice router, which can adaptively determine one token should continue to recurse or exit early \citep{rahmath2024early, yang2026dynamic}. Specifically, a router assigns each token a fixed recursion depth based on its initial hidden state, the token then unrolls the shared block for that many loops before exiting. To accelerate the autoregressive decoding, T‑LoopFormer employs recursion‑wise KV cache \citep{shi2024keep, li2024survey, cai2024pyramidkv}. Unlike approaches that retain key‑value pairs for all tokens across depths, this strategy selectively caches KV entries only for tokens that remain active at each recursion loop. As tokens progressively exit in deeper recursions, the cache size naturally shrinks. Moreover, attention is restricted exclusively to these locally cached entries, which can accelerate the decoding.

The effectiveness of our T‑Loopformer is thoroughly substantiated through extensive language modeling and zero-shot reasoning experiments. The dynamic token‑choice routing mechanism grants the model the flexibility to modulate the number of recursive steps according to each token's hidden state, departing from the uniform‑depth strategy that applies the same loop count indiscriminately, which decreases the accuracy of generated tokens. In addition, the recursion‑wise KV cache effectively accelerates the autoregressive decoding by only retraining the active tokens in the loop. The key contributions of this paper are summarized as follows:
\begin{itemize}
\item We propose T-LoopFormer, which can realize token-level elastic-depth through a lightweight dynamic token-choice router. Unlike fixed-depth models, this dynamic token-choice router can adaptively determines whether the token should exit early or continue recursing based on its hidden state, different tokens require different recurrent depths, which can improve the token generation accuracy.
\item We design a recursion-wise KV cache that selectively retains KV caches only for currently active tokens while discarding the KV caches of tokens that have exited the loop, thereby substantially accelerating autoregressive decoding of the model.
\item Extensive experiments on a wide range of language modeling and reasoning benchmarks demonstrate that T-LoopFormer can achieve robust performance while delivering superior decoding efficiency compared with baselines. Moreover, ablation studies validate the effectiveness of the token-choice router and recursion-wise KV cache. 
\end{itemize}

\section{Related Work}
\paragraph{Looped Transformers.}
Parameter sharing provides an orthogonal route to efficiency
and effective depth \citep{jeddi2026loopformer}. Through the introduction of adaptive computation time, the Universal Transformer establishes that repeatedly reusing a single block can achieve representational performance comparable to that of deep stacks without weight sharing \citep{dehghani2018universal}. During pretraining, ALBERT further reveals that extensive cross‑layer weight tying achieves considerable parameter efficiency, while downstream performance remains uncompromised \citep{lan2019albert}. Building upon this paradigm, Deep Equilibrium Models (DEQ) define an implicitly deep transformation with tied weights, solved iteratively via fixed‑point methods and implicit differentiation \citep{bai2019deep}. Besides, there are some works that explore looped transformers as programmable computers \citep{giannou2023looped} and mechanisms for algorithmic length generalization \citep{fan2025looped, jeddi2026loopformer}. What's more, Time‑step‑conditioned looping approaches include TMLT (Time-Modulated Looped Transformers) and LoopFormer, TMLT examine the representational capacity of looped Transformers in language modeling and demonstrate that conditioning on the timestep yields both improved scaling and lower perplexity \citep{xu2024expressive}. LoopFormer incorporates timestep and step‑size conditioning into looped Transformers, via shortcut‑consistency training over variable‑length trajectories, it enables elastic‑depth inference, where performance scales gracefully with the chosen compute budget \citep{jeddi2026loopformer}. However, these methods treat all tokens identically, forcing each of them to undergo the same fixed number of recursion steps regardless of their own hidden states, which may lead to a decrease in accuracy when generating tokens.
\paragraph{Latent Reasoning.}
looped Transformers possess an inductive bias for reasoning that strengthens with increasing effective computational depth and such abilities are framed as latent reasoning \citep{saunshi2024inductive, saunshi2025reasoning}. Unlike explicit chain‑of‑thought (CoT) prompting \citep{goyal2024think, cheng2024compressed, pfau2024let, kaissis2026step, chen2026loop, zhu2025scaling, jolicoeur2025less}, latent reasoning methods like looped transformers circumvent the need for verbalized reasoning steps, which significantly reduces the required context length and token consumption \citep{hao2024training, saunshi2025reasoning, jeddi2026loopformer}. Simultaneously, both theoretical and empirical researches link reasoning ability with network depth and algorithmic generalization \citep{merrill2024expressive}. Our method leverages the inductive bias inherent in looped Transformers and extends it to the token-level, enabling dynamic loop unrolling at the token-level latent reasoning. This allows the autoregressive decoding process to operate directly on the hidden states of each individual token.
\paragraph{Dynamic Routing.}
Dynamic routing enhances model efficiency by adapting computation to input complexity. Capsule Networks introduced routing-by-agreement for dynamic part-whole assignment without static pooling \citep{sabour2017dynamic}. In Transformers, ITT applies adaptive token-level routing with iterative hidden-state refinement, allowing critical tokens to undergo deeper computation while skipping redundant processing for simple tokens \citep{chen2025inner}. In MoE, difficulty-aware routing activates more experts for complex inputs and fewer for easy ones, overcoming fixed top-\(k\) selection \citep{huang2024harder}, while AdaMoE further enables token-adaptive allocation with null experts and differentiable end-to-end training \citep{zeng2024adamoe}. MoD \citep{raposo2024mixture} enables token-level dynamic compute in standard Transformers via top-\(k\) routing but its non-causal routing and discontinuous KV cache hinder parameter-tied looped architectures. In contrast to prior dynamic routing in standard Transformers, which selects among independent layers or experts, we extend dynamic routing to token-level loop unrolling within a parameter-tied looped Transformer, where each token's hidden state governs how many times the shared recurrent block is applied.
\paragraph{KV Cache.}
The KV cache has become a memory and bandwidth bottleneck for LLM inference as context lengths grow. \textbf{Eviction-based} methods selectively discard less critical KV pairs: CAKE \citep{qin2025cake} frames eviction as a cake-slicing problem that allocates per-layer cache sizes via spatial-temporal attention dynamics, achieving 10$\times$ speedup with 3.2\% cache; CriticalKV \citep{feng2025criticalkv} provides a perturbation-constrained selection algorithm that minimizes worst-case output perturbation, halving compression loss across benchmarks; ForesightKV \citep{dong2026foresightkv} learns to predict eviction via supervised training and RL, outperforming prior methods under half the cache budget. \textbf{Quantization-based} methods reduce bit-width: PolarQuant \citep{wu2026polarquant} uses polar transformation to handle key cache outliers, converting query-key inner products into table lookups for decoding acceleration; CommVQ \citep{li2025commvq} introduces commutative vector quantization with a RoPE-commutative codebook, achieving 87.5\% cache reduction with 2-bit quantization. In this work, we propose a recursion-wise KV cache that retains KV caches only for active tokens in the loop, which is an exact, loop-indexed cache management method tailored to recurrent unrolling, rather than an eviction heuristic or a lossy-approximation KV cache method such as those mentioned above.

\section{T-LoopFormer}
We introduce T-LoopFormer (see Figure \ref{fig:t_loopformer}), which is a looped transformer that utilizes a light weight dynamic token-choice router to achieve the token-level elastic-depth based on the hidden state of a token and use the recursion-wise KV cache during inference to accelerate autoregressive decoding. We use $X=(x_1,...,x_K)$ to denote a sequence of $K$ tokens which are drawn from a vocabulary $V$. Following LoopFormer \citep{jeddi2026loopformer}, the positional embeddings are added in a one-shot manner for simplicity we use the simplest cycle design, where a stack of $J$ Transformer layers denoted by $\Phi_{J} (\cdot)$. 
\subsection{Long and Short Trajectory Alignment}
We align the long and short trajectories produced during autoregressive 
decoding, ensuring that the model can flexibly adjust its recursion depth in 
the \textit{sequence-level}. The establishment of a sentence-level global 
inference loops provides a computational upper bound for token-level elastic 
depth recursion and the number of loops per token is determined by a 
\textbf{dynamic token-choice router}. Following the work 
\citep{jeddi2026loopformer}, T-LoopFormer uses a user-defined training loops $L$, the user specifies a step 
schedule $\Delta_L$ such that $\sum_{i=1}^{L} \Delta_i = 1$. T-LoopFormer then 
applies $\Phi_{J}(\cdot)$ for $L$ iterations, conditioning each loop $i$ on the 
cumulative time $c_{i-1}$ and step size $\Delta_i$, where 
$0 = c_0 < \cdots < c_L = 1$ and $\Delta_i = c_i - c_{i-1}$. The sequence 
$\Delta_L = (\Delta_1, \ldots, \Delta_L)$ is referred to as a 
\textit{trajectory}. The full 
trajectory corresponds to the maximum $L$ loops, each with $\Delta_i = 1/L$. 
Besides, T-LoopFormer is a decoder-only looped Transformer in which a single 
shared stack is applied iteratively. At iteration $l$, the model conditions on 
the pair $(c_{i-1}, \Delta_i)$, where $c_{i-1} \in [0,1]$ is the cumulative 
normalized time and $\Delta_i \in [0,1]$ is the step size. Encoded with 
sine--cosine frequency embeddings and projected via small MLPs, the two 
scalars yield $e_c$ and $e_{\Delta}$, and summing them produces 
$e_i = e_c + e_{\Delta}$. Modulation of the \textit{T-LoopFormer Block} is 
performed via this signal: an MLP maps $e_i$ to scaling $(\gamma_1, \gamma_2)$ 
for the two RMSNorm layers and to gating $(\alpha_1, \alpha_2)$ applied 
immediately before the residual connections of MHSA and FFN. T-LoopFormer 
uses shortcut conditioning together with an alignment loss 
$\mathcal{L}_{align}$ that drives trajectories of different lengths to align 
with the complete trajectory of length $L$. To sample the shortcut trajectory 
during training, we start from a T-LoopFormer with $L$ (unrolled $L$ times) and a maximum trajectory $\Delta_L$. In each 
batch, we first draw a shortcut length $S \sim \mathcal{U}\{1, \ldots, L-1\}$, 
and subsequently sample the step schedule $\Delta_S$ uniformly over $[0,1]$ 
subject to $\sum_{i=1}^{S} \Delta_i = 1$. The loss objective is:
\begin{equation}
\label{eq:1}
    \mathcal{L}=\mathcal{L}_{L}+\lambda_{1}\mathcal{L}_{S}
    +\lambda_{2}\mathcal{L}_{align},
\end{equation}
where $\mathcal{L}_{L}$ and $\mathcal{L}_{S}$ correspond to the next-token 
prediction losses for the longest and sampled shortcut trajectories, 
respectively, and $\mathcal{L}_{align}$ is a stop-gradient loss aligning per-token logits of shorter
trajectories to those of the longest trajectory. We set $\lambda_1=\lambda_2=0.1$ in all experiments.
\begin{figure}[htbp]
    \centering
    \begin{subfigure}[b]{0.49\textwidth}
        \centering
        \includegraphics[width=\textwidth, height=6cm]{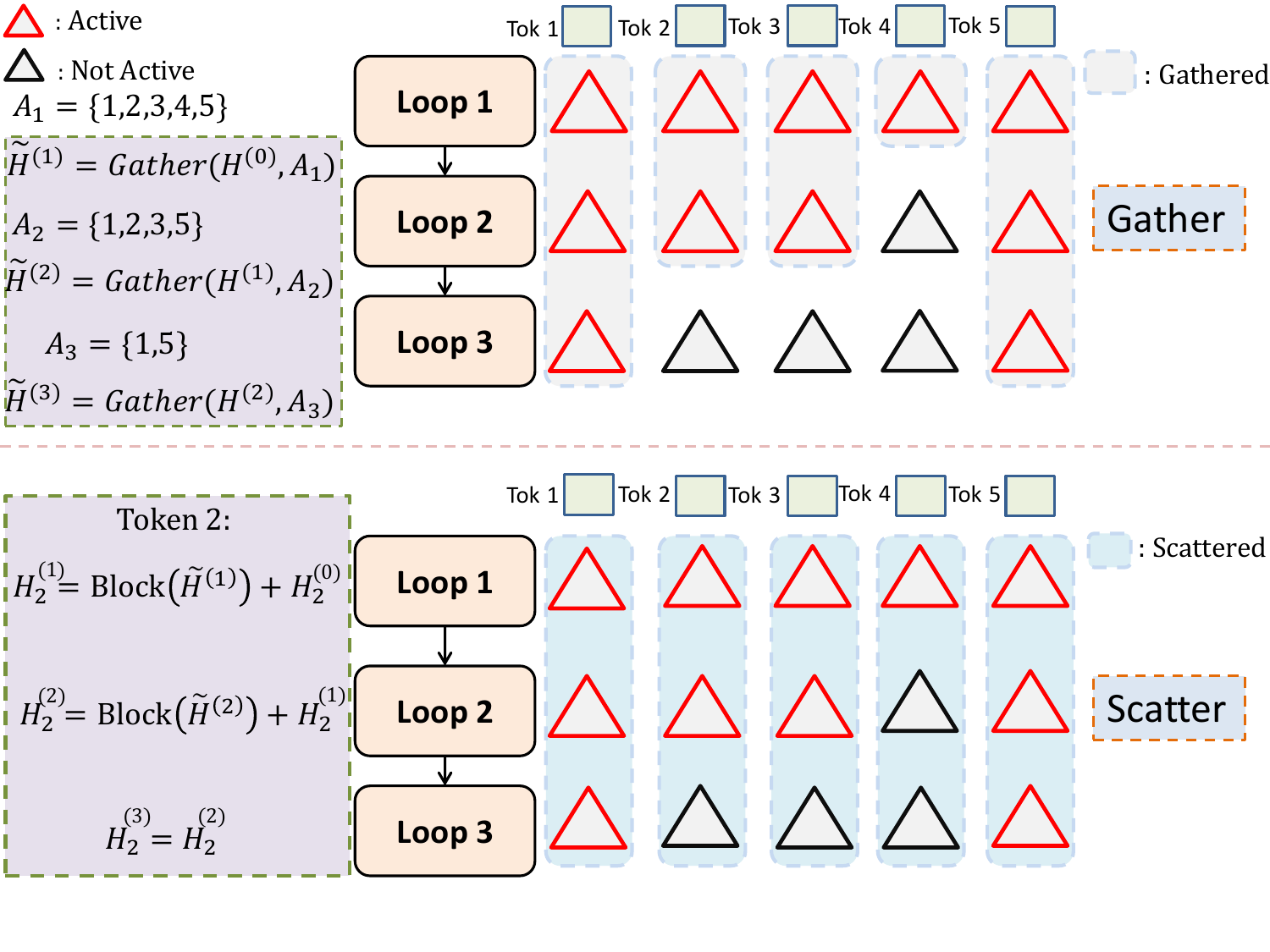}
        \caption{Gather-scatter during training.}
        \label{fig:training_gather_scatter}
    \end{subfigure}
    \hfill
    \begin{subfigure}[b]{0.49\textwidth}
        \centering
        \includegraphics[width=\textwidth,height=6cm]{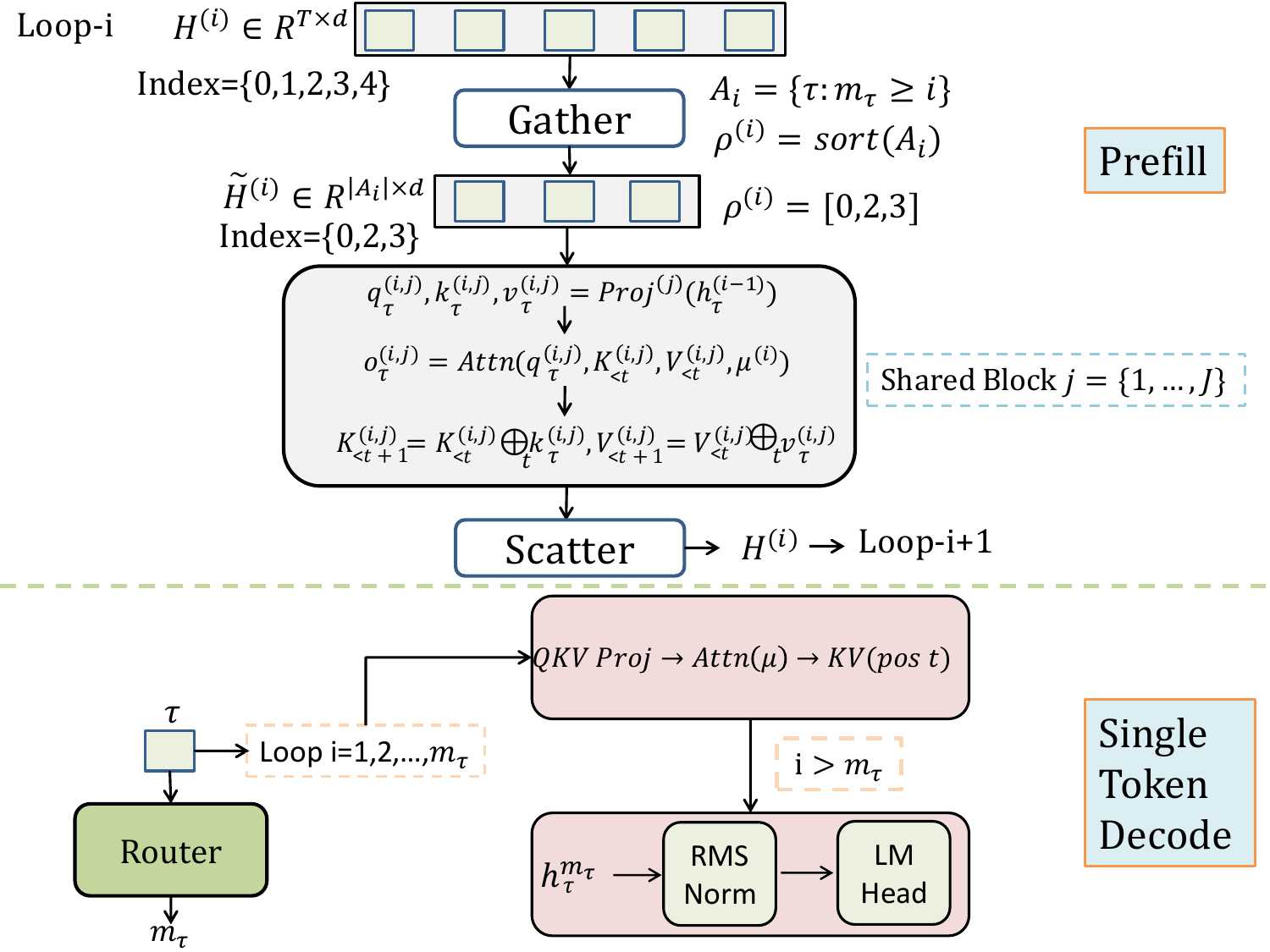}
        \caption{Prefill and decode during inference.}
        \label{fig:kv_cache}
    \end{subfigure}
    \caption{T-LoopFormer training and inference pipeline. (a): the gather-scatter during training; (b): prefill and decode during inference.}
    \label{fig:both}
\end{figure}
\subsection{Dynamic Token-Choice Routing}
The implementation of token-level elastic-depth unroll is based on 
sequence-level elastic-depth expansion. We use a two-layer MLP to serve as 
the lightweight dynamic token-choice router. Loops are indexed by 
$i \in \{1, \ldots, M\}$, where $M$ is the sequence-level maximum recursion 
depth shared by all tokens (during training, $M$ equals the sampled shortcut 
length $S$, or $L$ for the full trajectory; during inference, $M$ is set by 
the user). Let $h_{\tau}^{(i)} \in \mathbb{R}^{d}$ denote the 
hidden state of token $\tau$ after $i$ loops, with  the 
initial embedding $h_{\tau}^{(0)}$ and  ($h_{\tau}^{(0)}=E_{\tau}+E_{pos(\tau)} \in \mathbb{R}^{d}$); loop $i$ maps $h_{\tau}^{(i-1)}$ to $h_{\tau}^{(i)}$. The 
router $f_\theta: \mathbb{R}^d \to \mathbb{R}^M$ maps the initial hidden 
state to a distribution over recursion depths via a softmax layer:
\begin{equation}
\label{eq:2}
\pi_{\tau} = \text{softmax}\bigl(f_\theta(h_{\tau}^{(0)})\bigr) \in 
\mathbb{R}^{M},
\end{equation}
where $\pi_{\tau,\ell}$ is the probability that token $\tau$ requires depth 
$\ell$, $\ell \in \{1, \ldots, M\}$. For each loop $i$, the router computes 
the cumulative probability that token $\tau$ requires a recursion depth of 
at least $i$: 
\begin{equation}
\label{eq:3}
    p_{\tau}^{(i)} = \sum_{\ell=i}^{M} \pi_{\tau,\ell},
\end{equation}
and token $\tau$ executes loop $i$ (i.e., remains active) only if 
$p_{\tau}^{(i)}>0.5$:
\begin{equation}
\label{eq:4}
    \mathds{1}\bigl[\tau \in \mathcal{A}_{i}\bigr]
    =\mathds{1}\bigl[p_{\tau}^{(i)}>0.5\bigr],
\end{equation}
where $\mathcal{A}_{i} \subseteq \{ 1,\ldots,T\}$ denotes the set of tokens 
still active at loop $i$. Since $p_{\tau}^{(i)}$ is non-increasing in $i$, 
the active sets are automatically nested, 
$\mathcal{A}_{1} \supseteq \mathcal{A}_{2} \supseteq \cdots \supseteq 
\mathcal{A}_{M}$, and $\mathcal{A}_{1} = \{1,\ldots,T\}$ holds by 
construction as $p_{\tau}^{(1)} = 1$. A token is thus permanently 
deactivated once dropped (the routing decision is monotone over loops), and 
each token adaptively determines its own token-level recursion depth:
\begin{equation}
\label{eq:5}
    m_{\tau}=\max\bigl\{i\in \{1,\ldots,M \}:\tau\in \mathcal{A}_{i}
    \bigr\}, \qquad m_{\tau} \in \{1,\ldots,M\},
\end{equation}
where $m_{\tau}=1$ means the token traverses only a single pass of the 
shared block, whereas $m_{\tau}=M$ indicates the token propagates through 
all $M$ iterations. The trajectory is $\Delta_{m_{\tau}}=(\Delta_{m_{1}},...,\Delta_{m_{\tau}})$ and $\sum_{i=1}^{m_\tau} \Delta_i = 1$. Equivalently, $\tau \in \mathcal{A}_{i}$ if and only if 
$m_{\tau} \geq i$: the token's depth directly bounds the loops it 
participates in, with no index offset. As the routing distribution is 
computed once from the initial hidden state $h_{\tau}^{(0)}$, the per-token 
depth allocation is decided in a single forward evaluation of the 
lightweight router. Since the forward pass yields discrete values while backpropagation requires gradients, we employ the STE \citep{bengio2013estimating} to address this issue. See Appendix \ref{routing_discussion} for more routing discussions.

\paragraph{Gather--Scatter computation for training efficiency.} Because 
routing decisions vary per token, a naive implementation still executes 
the shared block over the full sequence at every loop, forfeiting any 
computational savings. We therefore couple the router with a gather-scatter 
mechanism that performs the lossless sequence compression during training. Let $H^{(i-1)} \in \mathbb{R}^{T \times d}$ collect the loop-$i$ input 
hidden states of all $T$ tokens. At each loop $i$, we first \textit{gather} the hidden states of active 
tokens into a compacted tensor:
\begin{equation}
\label{eq:6}
    \tilde{H}^{(i)} \;=\; \mathrm{Gather}\!\left(H^{(i-1)},\, \mathcal{A}_{i}
    \right) \;\in\; \mathbb{R}^{|\mathcal{A}_{i}| \times d},
\end{equation}
so that the shared block operates only on $\tilde{H}^{(i)}$, and the 
computation cost of loop $i$ scales with the number of surviving tokens 
$|\mathcal{A}_{i}|$ rather than the sequence length $T$. The updated states 
are then \textit{scattered} back to their original positions (this restores $H$ as a full-length tensor in the original sequence order, so that (i) the next loop can gather its own active set $\mathcal{A}_{i+1}$, which differs from $\mathcal{A}_i$; (ii) token $\tau$ in each position retains its latest state at a fixed coordinate for loss alignment; and (iii) gradients can flow back through the indexed assignment to the corresponding rows in the compact tensor):
\begin{equation}
\label{eq:7}
H_{\tau}^{(i)} = 
\begin{cases}
\mathrm{Block}\bigl(\tilde{H}^{(i)}\bigr) 
+ H_{\tau}^{(i-1)}, & {\tau} \in \mathcal{A}_{i}, \\[4pt]
H_{\tau}^{(i-1)}, & {\tau} \notin \mathcal{A}_{i}.
\end{cases}
\end{equation}
Tokens that exit the recursion simply retain the hidden 
state from their last executed loop, $h_{\tau}^{(m_{\tau})}$, which is 
passed to the final normalization and language-modeling head. 
\subsection{Recursion-wise KV Cache}
During inference, T-LoopFormer employs a recursion-wise KV cache mechanism 
to retain the KV cache for tokens in active loops while discarding the KV 
cache for inactive loops, thereby accelerating token generation. For each token, we allocate $M$ independent KV caches, one for each recursion loop 
$i \in \{1,\ldots,M\}$. Let $j \in \{1,\dots,J\}$ denote the layer index 
within the shared block, where $J$ is the total number of layers, $t^{'}$ are the historical keys positions. We define 
the cache state at recursion loop $i$, layer $j$, up to sequence position 
$t-1$ as
\begin{equation}
\label{eq:8}
K_{<t}^{(i,j)} = \bigl(k_{t'}^{(i,j)}\bigr)_{t' < t}, 
\qquad
V_{<t}^{(i,j)} = \bigl(v_{t'}^{(i,j)}\bigr)_{t' < t}.
\end{equation}

Crucially, inference is also executed with the $gather-scatter$ scheme (see Appendix \ref{gather_scatter_diff}). The active rows are first gathered into a 
compact tensor $\tilde{H}^{(i)}$ (\eqref{eq:6}), where the gather indices are sorted so that the causal order is preserved 
within the compact sequence. Only $\tilde{H}^{(i)}$ is passed through the 
shared block. Importantly, the original sequence positions 
$\rho^{(i)} = \mathrm{sort}(\mathcal{A}_{i})$ are captured at gather time 
and propagated through the block to the cache-writing routine: within each 
attention layer, the newly computed keys and values are written into the 
loop-$i$ cache at their original positions $\rho^{(i)}$ rather than at 
their compact indices. Positional bookkeeping maintains causal alignment via original-position cache references. After the block 
returns the updated compact states, they are scattered 
back to their original positions and get $H^{(i)}$ (\eqref{eq:7}),
where $\mathrm{scatter}$ overwrites only the rows indexed by 
$\mathcal{A}_{i}$; the hidden states of exited tokens are left untouched. In the single-token decoding case, the scheme 
degenerates to an early exit: the recursion loop terminates as soon as the 
current token leaves the active set.

During autoregressive decoding of token $\tau$ (arriving at sequence 
position $t$, so that $K_{<t}$ holds exactly the tokens preceding it), its 
routing depth $m_{\tau}$ is determined by the router. For 
each recursion loop $i$, the token computes its 
query, key, and value projections at layer $j$ from its incoming state:
\begin{equation}
\label{eq:9}
q_{\tau}^{(i,j)},\, k_{\tau}^{(i,j)},\, v_{\tau}^{(i,j)} = 
\mathrm{Proj}^{(j)}\bigl(h_{\tau}^{(i-1)}\bigr).
\end{equation}
The attention mechanism queries only the historical keys and values 
belonging to the same loop $i$, restricted to the valid entries of that 
cache:
\begin{equation}
\label{eq:10}
o_{\tau}^{(i,j)} = \mathrm{Attn}\bigl(q_{\tau}^{(i,j)},\, K_{<t}^{(i,j)},\, 
V_{<t}^{(i,j)};\, \mu^{(i)}\bigr),
\end{equation}
where the validity indicator $\mu^{(i)}_{t^{'}} = \mathds{1}[\,m_{t^{'}} \geq i\,]$ 
excludes the cache slots of tokens that exited before loop $i$ (in batched 
prefill, a block-causal mask is additionally applied among the currently 
gathered tokens). Masking is required to avoid softmax assigning non‑negligible mass to zero slots of exited tokens.

Whenever token $\tau$ executes loop $i$ (i.e., $m_{\tau} \geq i$), its 
current key and value must be stored, since future tokens that also reach 
loop $i$ will attend to them. We update the loop-specific cache by writing 
the new states at the original sequence position:
\begin{equation}
\label{eq:11}
K_{<t+1}^{(i,j)} = K_{<t}^{(i,j)} \oplus_{t} k_{\tau}^{(i,j)}, \qquad
V_{<t+1}^{(i,j)} = V_{<t}^{(i,j)} \oplus_{t} v_{\tau}^{(i,j)},
\end{equation}
where $\oplus_{t}$ denotes writing the new state at cache position $t$. For 
single-token decoding, this reduces to ordinary sequence concatenation at 
the tail of the cache; for batched prefill, the writes are indexed by the 
gathered positions $\rho^{(i)}$ of all active tokens. If the token exits 
the recursion after loop $m_{\tau}$ (i.e., for loops $i > m_{\tau}$), it 
neither computes new projections nor updates any cache. Its final hidden 
state $h_{\tau}^{(m_{\tau})}$ is directly passed to the final normalization 
and LM head, bypassing the remaining shared block iterations.

\begin{table}[ht]
\tiny
\setlength{\tabcolsep}{2.0pt}  
\renewcommand{\arraystretch}{0.9}  
\centering
\caption{The perplexity comparison on the FineWeb‑Edu and OpenWebText validation sets and the accuracy comparison on ten zero‑shot reasoning tasks between our model and all baseline models under various FLOPs. The best performance is marked in \textbf{bold}, the second‑best is \underline{underlined}, and our model's results are highlighted.}
\label{tab:main}
\begin{tabular}{l|c|cc|cccccccccc|c}
\toprule
\multirow{2}{*}{} & \multirow{1}{*}{Params} & \multicolumn{2}{c|}{Perplexity $\downarrow$} & \multicolumn{10}{c}{Language Tasks (Accuracy) $\uparrow$} & \multirow{2}{*}{Avg Acc} \\
\cline{3-4} \cline{5-14} & & FineWeb-Edu & OpenWebText & COPA & HS & LB & OBQA & PIQA & Race & SciQ & ARC-C & SIQA & ARC-E & \\
\midrule
\multicolumn{14}{l}{\textbf{FLOPs: 24x}} \\
\midrule
Base (24 $\otimes$ 1) & 24x  & \underline{20.17} & \underline{19.22} & 62 & \textbf{34.79} & \textbf{41.98} & 27.57 & \underline{66.29} & 29.78 & \textbf{70.22} & \underline{27.79} & \underline{38.83} & \underline{38.76} & \underline{43.8} \\
Base-Loop (3 $\otimes$ 8) & 3x  & 23.23 & 22.78 & 61 & 31.65 & 35.83 & 27.11 & 64.45 & 28.93 & 63.79 & 25.11 & \textbf{38.94} & 36.12 & 41.29 \\
TMLT (3 $\otimes$ 8) & 3x & 22.87 & 21.33 & 66 & 33.22 & 40.06 & \underline{27.92} & 64.11 & 30.14 & \underline{70.11} & 25.55 & 37.32 & 36.58 & 43.1 \\
Naive-Loop-EE (3 $\otimes$ 8) & 3x  & 25.54 & 24.23 & \underline{67} & 30.78 & 31.89 & 27.03 & 63.42 & 29.56 & 66.32 & 25.88 & 36.97 & 36.77 & 41.56 \\
Base-Loop-EE-Align (3 $\otimes$ 8) & 3x  & 24.21 & 23.25 & \underline{67} & 31.46 & 32.92 & 26.9 & 63.23 & 29.43 & 65.29 & 26.77 & 37.81 & 37.83 & 41.86 \\
TMLT-EE (3 $\otimes$ 8) & 3x  & 23.47 & 21.98 & \underline{67} & 32.44 & 36.92 & \textbf{28.4} & 64.14 & 28.9 & 68.1 & 26.68 & 37.73 & 37.66 & 42.80 \\
LoopFormer(3 $\otimes$ 8) & 3x  & 21.96 & 20.09 & \textbf{68} & 32.93 & 39.32 & 26.7 & 64.95 & \underline{31.19} & 68.44 & 27.56 & 38.19 & 37.49 & 43.48 \\
\rowcolor{yellow!20}
T-LoopFormer(3 $\otimes$ 8) & 3x  & \textbf{20.13} & \textbf{19.14} & \textbf{68} & \underline{33.67} & \underline{40.18} & 27.33 & \textbf{66.32} & \textbf{31.88} & 68.76 & \textbf{28.04} & 38.68 & \textbf{39.17} & \textbf{44.20} \\
\midrule
\multicolumn{14}{l}{\textbf{FLOPs: 12x}} \\
\midrule
Base (12 $\otimes$ 1) & 12x  & \textbf{21.69} & \textbf{21.11} & \underline{68} & \textbf{32.82} & \textbf{37.78} & \textbf{26.28} & \underline{64.69} & \textbf{29.88} & \textbf{69.21} & \underline{26.31} & \underline{38.41} & \underline{38.17} & \textbf{43.16} \\
Naive-Loop-EE (3 $\otimes$ 4) & 3x  & 24.79 & 25.12 & 66 & 29.67 & 31.69 & 26.19 & 61.95 & 28.53 & 64.56 & 25.23 & 37.12 & 36.24 & 40.72 \\
Base-Loop-EE-Align (3 $\otimes$ 4) & 3x  & 26.01 & 25.83 & 63 & 29.98 & 26.07 & 27.43 & 61.63 & 28.26 & 58.89 & 25.25 & 36.35 & 36.74 & 39.36 \\
TMLT-EE (3 $\otimes$ 4) & 3x  & 26.36 & 26.77 & 62 & 30.6 & 27.94 & 26.79 & 62.31 & \underline{28.85} & 61.74 & 25.72 & 36.78 & 36.98 & 39.97 \\
LoopFormer(3 $\otimes$ 4) & 3x  & 24.23 & 23.69 & \underline{68} & 31.55 & 32.54 & 25.68 & 63.99 & 28.51 & 66.73 & 26.28 & 37.82 & 37.14 & 41.82 \\
\rowcolor{yellow!20}
T-LoopFormer(3 $\otimes$ 4) & 3x  & \underline{23.6} & \underline{23.14} & \textbf{69} & \underline{32.33} & \underline{33.26} & \underline{26.19} & \textbf{64.82} & 28.79 & \underline{67.11} & \textbf{26.96} & \textbf{38.47} & \textbf{38.85} & \underline{42.58} \\
\midrule
\multicolumn{14}{l}{\textbf{FLOPs: 6x}} \\
\midrule
Base (6 $\otimes$ 1) & 6x  & \textbf{24.84} & \textbf{24.04}& \textbf{64} & \textbf{30.58} & \textbf{33.78} & \underline{25.78} & \textbf{62.58} & \textbf{28.31} & \textbf{67.61} & \textbf{31.16} & \underline{36.21} & \underline{37.66} & \textbf{41.77} \\
Naive-Loop-EE (3 $\otimes$ 2) & 3x & \underline{29.03} & \underline{28.12} & \underline{63} & 28.86 & \underline{27.51} & 24.82 & \underline{61.78} & 26.49 & \underline{62.38} & 30.02 & 35.53 & 36.13 & 39.65 \\
Base-Loop-EE-Align (3 $\otimes$ 2) & 3x & 35.07 & 34.32 & 60 & 28.33 & 18.88 & 25.67 & 59.88 & 25.24 & 54.02 & 29.05 & 34.79 & 36.23 & 37.21 \\
TMLT-EE (3 $\otimes$ 2) & 3x & 37.31 & 37.22 & 59 & 28.34 & 17.45 & 25.37 & 59.11 & 26.51 & 50.41 & 28.54 & 35.16 & 36.53 & 36.64  \\
LoopFormer (3 $\otimes$ 2) & 3x & 32.89 & 32.05 & \underline{63} & 28.81 & 26.68 & 26.87 & 60.67 & 26.03 & 59.02 & 29.67 & 35.77 & 37.02 & 39.35 \\
\rowcolor{yellow!20}
T-LoopFormer (3 $\otimes$ 2) & 3x & 32.03 & 31.76 & \textbf{64} & \underline{29.17} & 27.03 & \textbf{27.35} & 61.13 & \underline{26.5} & 59.64 & \underline{30.44} & \textbf{36.46} & \textbf{38.18} & \underline{39.99} \\
\bottomrule
\end{tabular}
\end{table}

\begin{table}[t]
    \centering
    \begin{minipage}{0.49\textwidth}
    \setlength{\tabcolsep}{1.0pt}  
        \centering
        \tiny
        \captionof{table}{TPOT of different models on FineWeb-Edu validation set at 24$\times$ FLOPs.}
        \label{tab:infer_latency}
        \begin{tabular}{lc}
            \toprule
            Model & Latency (ms/token) \\
            \midrule
            Base ($24 \otimes 1$) & \underline{0.202} \\
            Base-Loop (3 $\otimes$ 8) & 0.497 \\
            TMLT (3 $\otimes$ 8) & 0.389 \\
            Naive-Loop-EE (3 $\otimes$ 8) & 0.433 \\
            Base-Loop-EE-Align (3 $\otimes$ 8) & 0.425 \\
            TMLT-EE (3 $\otimes$ 8) & 0.312 \\
            LoopFormer (3 $\otimes$ 8) & 0.441 \\
            T-LoopFormer ($3 \otimes 8$) & \textbf{0.197} \\
            \bottomrule
        \end{tabular}
    \end{minipage}
    \hfill
    \begin{minipage}{0.49\textwidth}
    \setlength{\tabcolsep}{1.0pt}  
        \centering
        \tiny
        \captionof{table}{Training memory of different models on FineWeb-Edu training set at 24$\times$ FLOPs}
        \label{tab:train_mem}
        \begin{tabular}{lc}
            \toprule
            Model & Memory (MB) \\
            \midrule
            Base ($24 \otimes 1$) & 634,225 \\
            Base-Loop (3 $\otimes$ 8) & 93,442 \\
            TMLT (3 $\otimes$ 8) & 96,594 \\
            Naive-Loop-EE (3 $\otimes$ 8) & \underline{87,997} \\
            Base-Loop-EE-Align (3 $\otimes$ 8) & 91,256 \\
            TMLT-EE (3 $\otimes$ 8) & 98,387 \\
            LoopFormer (3 $\otimes$ 8) & 97,739 \\
            T-LoopFormer ($3 \otimes 8$) & \textbf{74,889} \\
            \bottomrule
        \end{tabular}
    \end{minipage}
\end{table}

\section{Experiments}
\label{experiments}
Following \citep{jeddi2026loopformer}, we compare a 24-layer, $\sim$1B-parameter non-looped Transformer with FLOP-matched looped variants. All models use a GPT-style decoder \citep{radford2019language} with NanoGPT configurations. Training is performed on FineWeb-Edu \citep{penedo2024fineweb} for 25B tokens in accordance with Chinchilla scaling \citep{hoffmann2022training}. See Appendix \ref{implementation} for details. Following \citep{jeddi2026loopformer}, in terms of model parameters, 24$\times$ indicates that the model contains 24 layers. In terms of FLOPS, 24$\times$ denotes the product of the number of model layers $J$ and the user-specified inference depth $M$, $FLOPs \propto J \otimes M$, ignoring embedding/unembedding costs. The token-choice router threshold we set is 0.5.
\paragraph{Evaluation Metrics and Benchmarks.}
Following \citep{saunshi2025reasoning, geiping2026scaling, jeddi2026loopformer}, we report perplexity and downstream zero-shot accuracy. Perplexity is measured on FineWeb-Edu \citep{penedo2024fineweb} and OpenWebText \citep{Gokaslan2019OpenWeb}. For latent reasoning, we report zero-shot accuracy on ten established benchmarks spanning a range of reasoning difficulty: COPA \citep{roemmele2011choice}, HellaSwag (HS) \citep{zellers2019hellaswag}, LAM-BADA 
(LB) \citep{paperno2016lambada}, OpenBookQA (OBQA) \citep{mihaylov2018can}, RACE \citep{lai2017race}, Social IQA (SIQA) \citep{sap2019social}, ARC-Easy (ARC-E), ARC-Challenge (ARC-C) \citep{clark2018think}, and
SciQ \citep{welbl2017crowdsourcing}.
\paragraph{Baselines.}
We compare T-LoopFormer against two groups of baselines: fixed-depth models and depth-elastic models. Fixed-depth models include Base: a non-looped Transformer; Base-Loop: a standard looped model; TMLT: a looped model with timestep conditioning. Depth-elastic models include Base-Loop-EE: naive early exiting applied to the basic looped model; TMLT-EE: early exiting and long-short trajectory alignment training applied to TMLT to enable depth elasticity; Base-Loop-EE-Align: augmented with the long-short trajectory alignment during training. LoopFormer: an elastic-depth looped transformers with long-short trajectory alignment. 
\begin{figure}[htbp]
    \centering
    \begin{subfigure}[b]{0.48\textwidth}
        \centering
        \includegraphics[width=\textwidth, height=4cm]{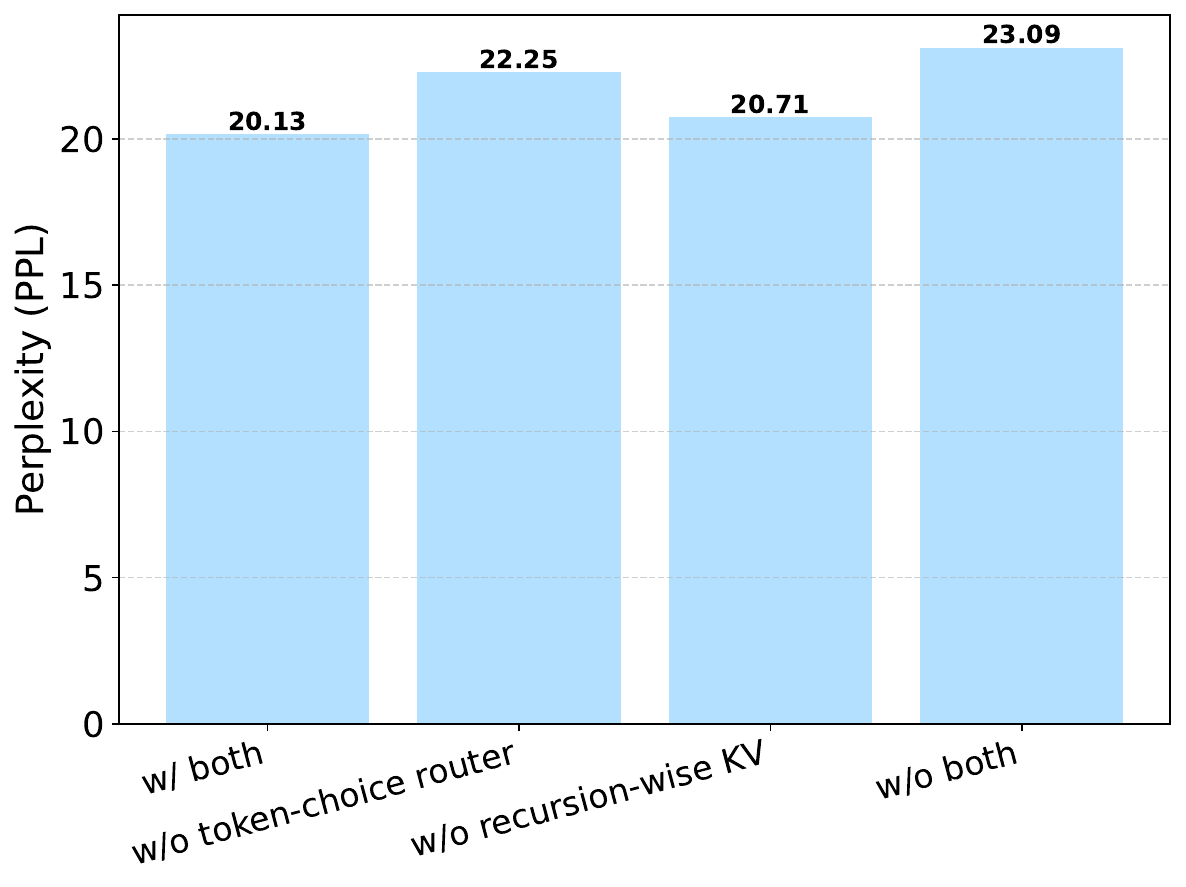}
        \caption{Perplexity (PPL)}
        \label{ablation_router_kv_ppl}
    \end{subfigure}
    \hfill
    \begin{subfigure}[b]{0.48\textwidth}
        \centering
        \includegraphics[width=\textwidth, height=4cm]{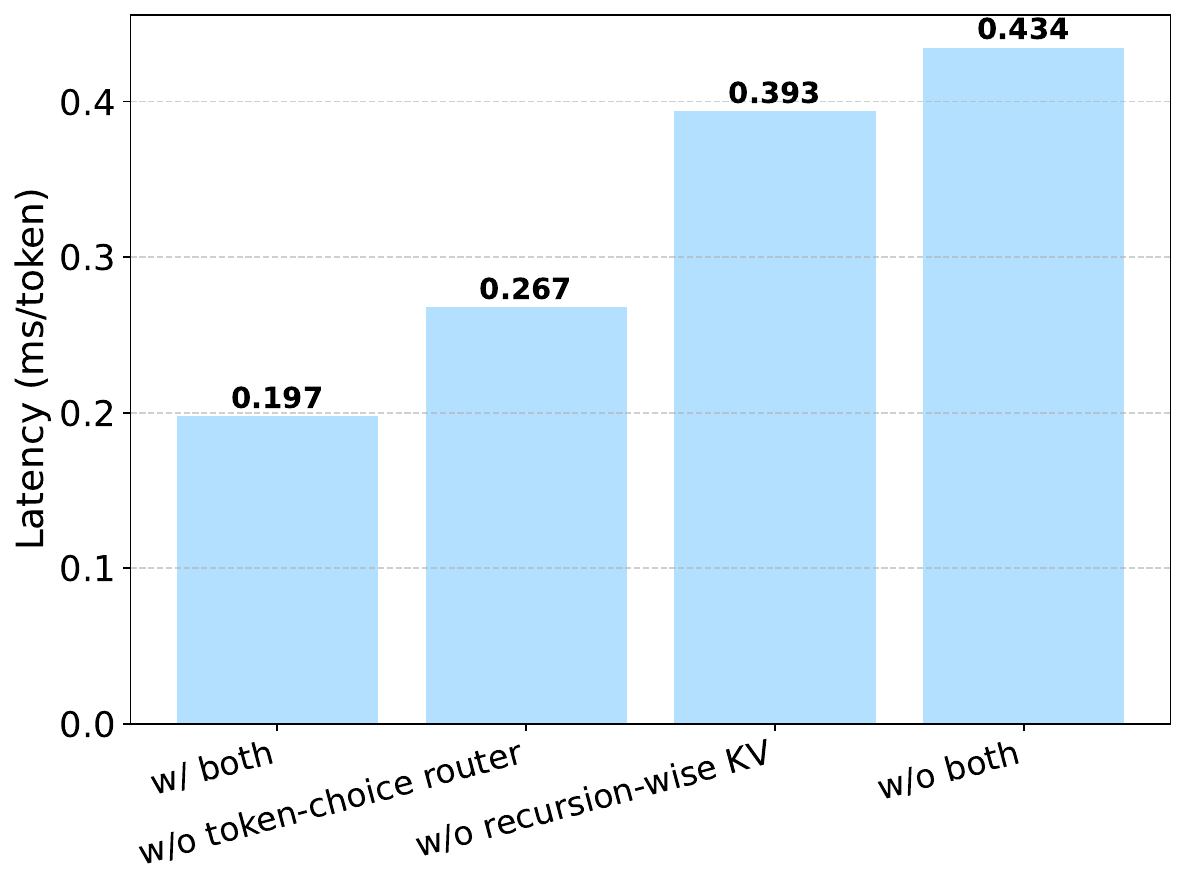}
        \caption{TPOT}
        \label{ablation_router_kv_latency}
    \end{subfigure}
    \caption{The influence of token-choice router and recursion-wise KV cache. Tested on FineWeb-Edu validation set at 24$\times$ FLOPs. (a) PPL, (b) TPOT.}
    \label{ablation_combined}
\end{figure}
\begin{figure}[htbp]
    \centering
    \begin{subfigure}[b]{0.48\textwidth}
        \centering
        \includegraphics[width=\textwidth, height=4cm]{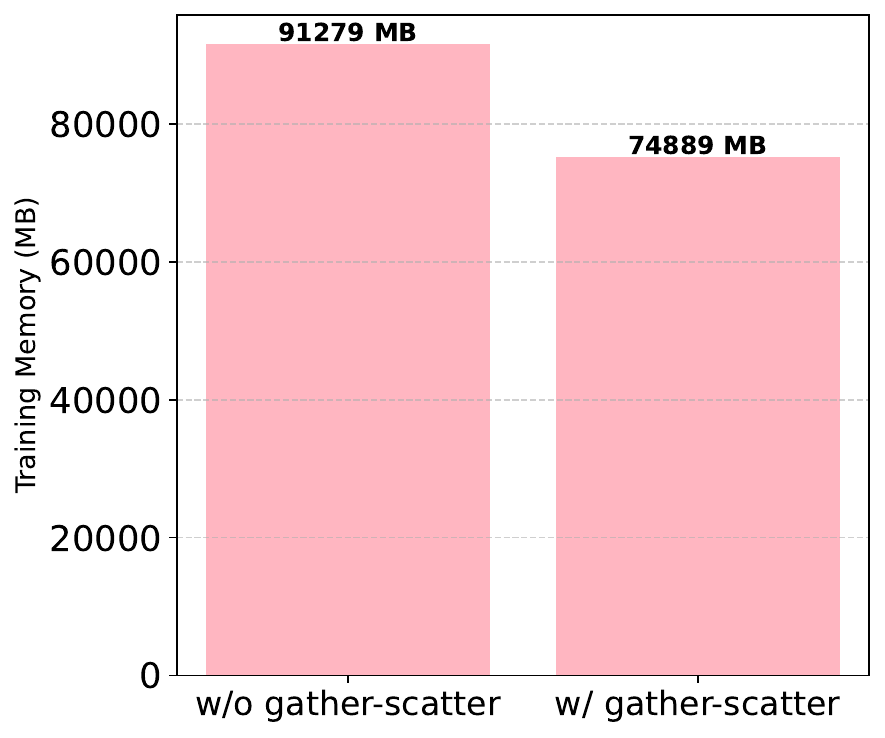}
        \caption{Training memory with/without gather-scatter.}
        \label{ablation_gather_scatter_train_mem}
    \end{subfigure}
    \hfill
    \begin{subfigure}[b]{0.48\textwidth}
        \centering
        \includegraphics[width=\textwidth, height=4cm]{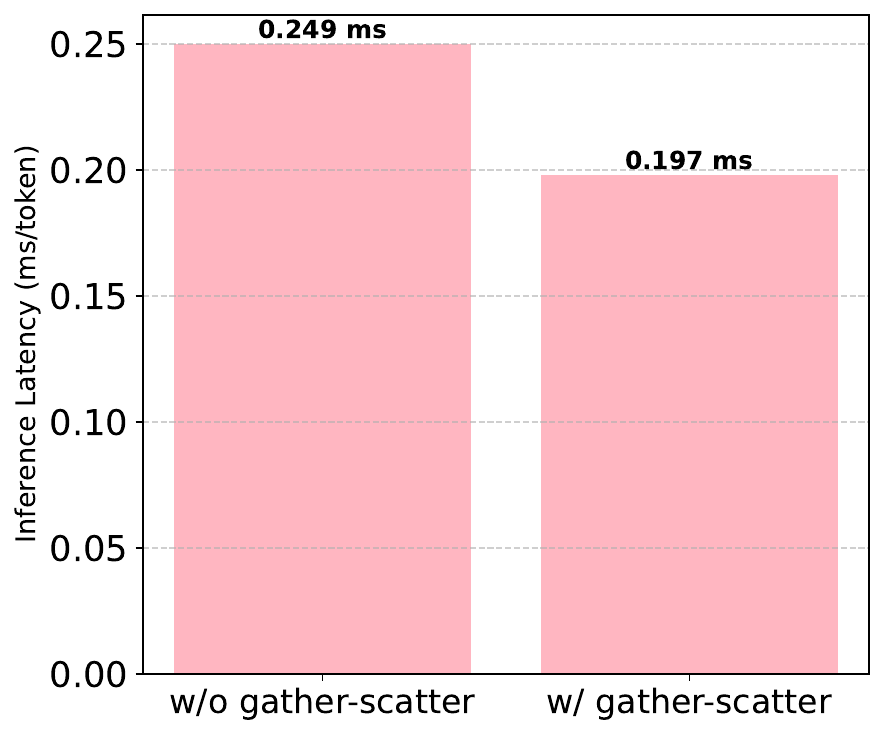}
        \caption{TPOT with/without gather-scatter.}
        \label{ablation_gather_scatter_latency}
    \end{subfigure}
    \\[1ex]  
    \begin{subfigure}[b]{0.48\textwidth}
        \centering
        \includegraphics[width=\textwidth, height=4cm]{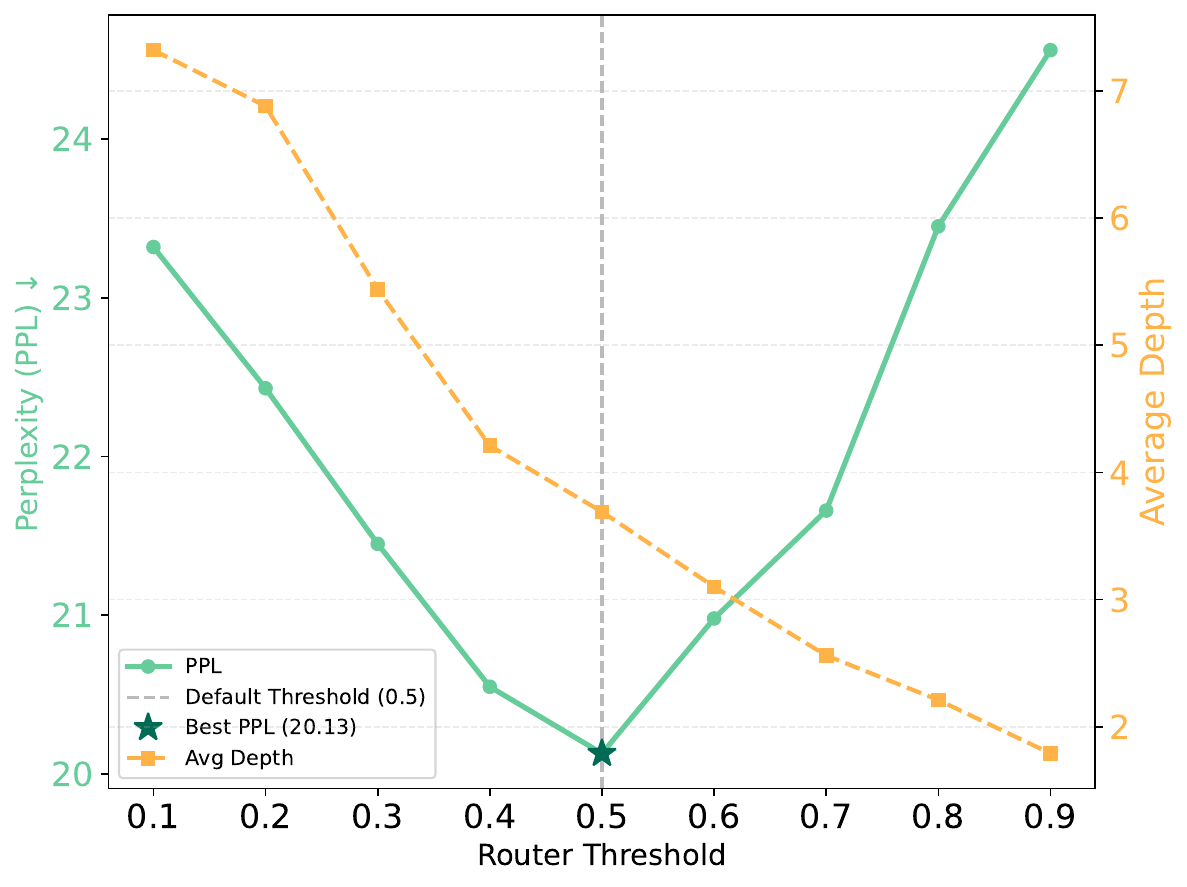}
        \caption{Perplexity and average depth under different router thresholds.}
        \label{ablation_router_threshold}
    \end{subfigure}
    \hfill
    \begin{subfigure}[b]{0.48\textwidth}
        \centering
        \includegraphics[width=\textwidth, height=4cm]{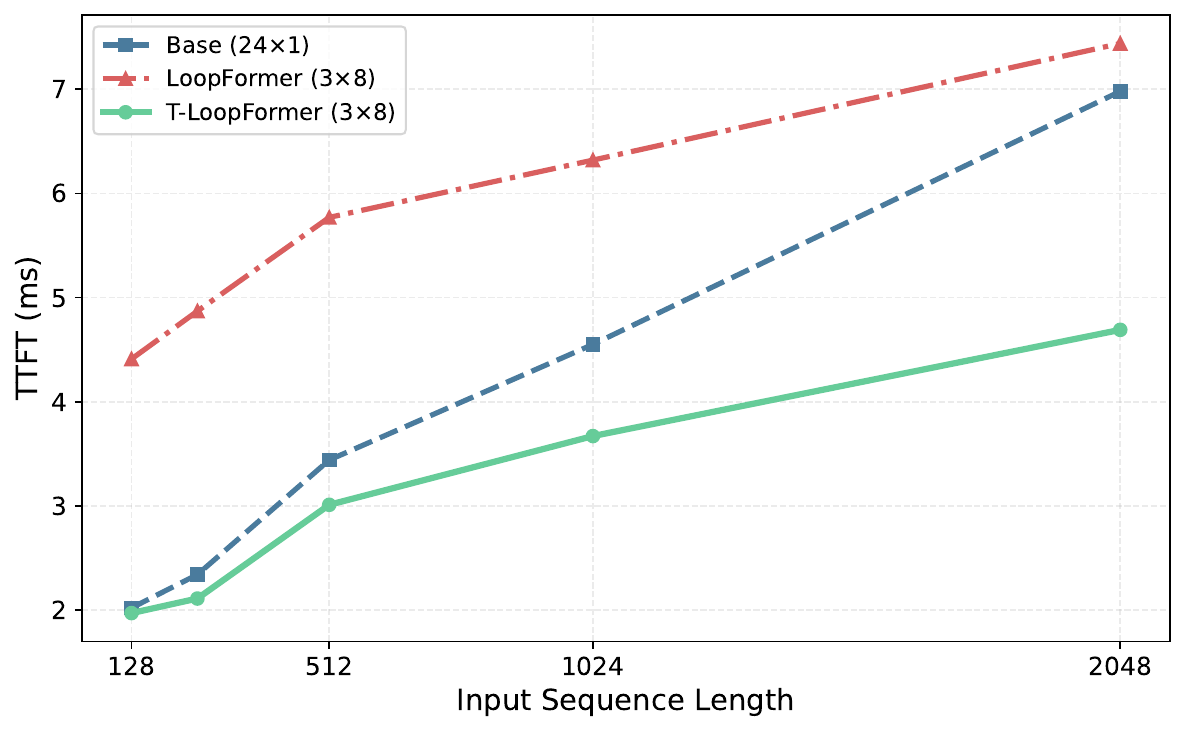}
        \caption{TTFT with different input sequence lengths.}
        \label{ablation_sequence}
    \end{subfigure}
    \caption{Ablation studies on key components of T-LoopFormer. (a) and (b): effect of the gather–scatter mechanism on training memory and inference latency; (c): sensitivity analysis of the router threshold; (d): scaling of TTFT with input sequence length.}
    \label{ablation_combined2}
\end{figure}
\subsection{Main Results}
Table \ref{tab:main} shows that, under different FLOPs, T-LoopFormer achieves the best reasoning performance among models with the same parameter. Moreover, at 24$\times$ FLOPs, our model outperforms the non-looped base model with 24$\times$ parameter. This demonstrates that the dynamic token-choice routing mechanism designed in our model enables each token to adaptively unroll different number of loops based on its hidden state, and that either excessive or insufficient elastic-depth unrolling may lead to incorrect token outputs. Under 12x and 6x FLOPs, because the maximum number of loops is reduced, some tokens require more loop to unroll, which causes those tokens to be insufficiently reasoned. As a result, T‑LoopFormer underperforms the non‑looped base model on perplexity and zero‑shot reasoning tasks. However, compared with other looped models, it narrows the performance gap with the non‑looped base model (see Appendix \ref{performance_degeneration} for more discussions). In addition to the comparison of model performance, Table \ref{tab:infer_latency} shows that our model achieves the lowest decoding latency among all baselines, even lower than the non‑looped base model, demonstrating that our model can improve single token decoding efficiency. Moreover, compared with LoopFormer, our model achieves 2.24$\times$ decoding speed faster than LoopFormer, which confirms that the dynamic token‑choice routing mechanism and the recursion‑wise KV cache are effective in improving inference speed. Table \ref{tab:train_mem} validates that T‑LoopFormer can achieve the lowest memory usage during training, indicating that it can improve the training efficiency. Figure \ref{fig:loop_ratio_pie} shows the distribution of loop counts per token on the FineWeb‑Edu validation set and most tokens require at most 4 loops (average depth is 3.69), and the overall distribution exhibits a long‑tail pattern. This indicates that different tokens need different numbers of loops, validating the effectiveness of token‑choice router.

\subsection{Ablation Study}

\paragraph{Influence of Token-Choice Router and Recursion-wise KV Cache.} 
Figure \ref{ablation_router_kv_ppl} exhibits that the token-choice router contributes more to generating correct tokens (without router, PPL increases 2.12, without recursion-wise KV cache, PPL increases 0.58) and figure \ref{ablation_router_kv_latency} validates that the recursion-wise  KV cache contributes more to increasing the decoding speed (without recursion-wise KV cache, TPOT increases 0.196 ms and without router, TPOT increases 0.70 ms).  

\paragraph{Influence of Gather-Scatter during Training and Inference.} 
Figure \ref{ablation_gather_scatter_train_mem} shows that without gather-scatter during training, memory increases 21.9\%, which validates that the gather-scatter mechanism can improve the training efficiency and figure \ref{ablation_gather_scatter_latency} shows that without gather-scatter during inference, the inference latency increases 26.4\% (tested on FineWeb-Edu validation set). These validate that gather-scatter mechanism can improve the autoregressive decoding efficiency.
\paragraph{Influence of Different Router Threshold.}
Figure \ref{ablation_router_threshold} shows the influence of different router threshold (tested on FineWeb-Edu validation set at 24x FLOPs), 0.5 is the best threshold and the perplexity (PPL) of the model equipped with the token‑choice router on the FineWeb‑Edu validation set exhibits a U‑shaped curve as the threshold varies.
\paragraph{TTFT with Different Input Sequence Lengths.}
Figure \ref{ablation_sequence} shows the TTFT (time to first token) with different input sequence lengths (128, 512, 1024, and 2048), we randomly sample 1,000 documents from FineWeb-Edu validation set, each with a raw length exceeding 2,048 tokens. We then truncate each to target lengths {128, 512, 1024, 2048} as the input prompt. The results show that, as input length grows, T‑LoopFormer shows a smaller increase in TTFT than the base model and achieves the lowest latency across all sequence lengths, demonstrating its efficient scaling with input length.
\section{Conclusion}
We build \textbf{T-LoopFormer}, a looped Transformer with token‑level elastic-depth for latent reasoning. A dynamic token-choice router adaptively assigns each token a variable number of recursion loops, allowing different tokens receive different computations. For efficiency, we incorporate a gather–scatter mechanism during training and inference, which restricts each loop to active tokens, reducing training memory and decoding latency, and a recursion‑wise KV cache that discards caches of exited tokens and keeps KV caches of active tokens in each loop to further accelerate autoregressive decoding. Together, these designs yield an efficient and flexible architecture. Experiments show that T-LoopFormer reaches the robust performance under the same parameters on language modeling and zero-shot reasoning tasks (even surpassing base Transformer at 24× FLOPs) and can reach the lowest decoding latency, which validates the effectiveness of our model. Future work will extend T‑LoopFormer to larger‑scale pretraining and broader reasoning tasks.
\section{AI Use Statement}
During the preparation of this manuscript, the authors used DeepSeek solely for language polishing and for assisting with data preprocessing, including drafting or refactoring preprocessing code and checking data-cleaning scripts. The tool was not used to generate scientific ideas, formulate methods, conduct experiments, interpret results, or write the core scientific claims. All AI-assisted outputs were carefully reviewed, verified, and edited by the authors, who take full responsibility for the content, integrity, and reproducibility of this work. The final data preprocessing was carried out by deterministic scripts.
\section{Reproducibility Statement}
To support reproducibility, we provide a complete description of the proposed method, experimental setup, and evaluation protocol in Section \ref{experiments} and Appendix \ref{implementation}. The source code, hyperparameters, random seeds, and step-by-step instructions for reproducing the main results are available in the anonymous supplementary repository at \url{https://anonymous.4open.science/r/T-LoopFormer-53BE}. We believe these resources are sufficient for reproducing the results reported in this paper.
\bibliography{iclr2027_conference}
\bibliographystyle{iclr2027_conference}

\clearpage

\appendix

\section{Implementation Details}
\label{implementation}
\paragraph{Hardware and framework.} All models are trained on 8$\times$A100 (80 GB) GPUs using the open-source NanoGPT training stack as a reference implementation.
\paragraph{Data and tokens.}Unless otherwise specified, we run each experiment for 250,000 optimizer steps
with a global batch size of 48 sequences and block size 2048 (context length). This corresponds to
approximately 25B training tokens in total.
\paragraph{Optimization.}We use AdamW with weight decay $2 \times 10^{-1}$
, cosine learning-rate decay (per
NanoGPT), peak learning rate $lr=6 \times 10^{-4}$
, minimum learning rate $min\_lr=6\times 10^{-5}$,
and 4,000 warmup steps, which we found important for stability. Similar to observations in \citep{geiping2026scaling} , we occasionally observe training instabilities at depth; warmup and cosine decay
mitigate these in practice. Unless noted, other optimizer and training defaults follow NanoGPT.
\paragraph{Model hyperparameters.}Following \citep{saunshi2025reasoning, jeddi2026loopformer}, we use hidden size d = 2048 and
nheads = 32 for all configurations. The feed-forward dimension is $d_{ff}$ = 5120 with a standard
two-layer GELU MLP. All normalizations are RMSNorm. We use learned positional embeddings
added to token embeddings (NanoGPT default). 
\paragraph{T-LoopFormer Conditioning.}
We use two embedding modules, one for normalized time \(t \in [0,1]\) and one for step size \(\Delta t \in (0,1]\). Each maps a scalar input to a \(d\)-dimensional conditioning vector via (i) fixed sinusoidal Fourier features (width \(D_f = 256\), max period 10000), followed by (ii) a 2-layer MLP with hidden size \(d\) and SiLU activation:
\[
\phi(\tau) = \mathrm{MLP}\left(
\left[
\cos(\tau \omega_1), \sin(\tau \omega_1), \ldots,
\cos(\tau \omega_{D_f/2}), \sin(\tau \omega_{D_f/2})
\right]
\right) \in \mathbb{R}^d
\]
where \(\omega_k = \exp\left( -\frac{k-1}{D_f/2} \log 10{,}000 \right)\) for \(k=1,\ldots,D_f/2\). Given per-token cumulative time \(t_{\tau,i}\) and step size \(\Delta t_{\tau,i}\) for each active token \(\tau\) at loop \(i\), we compute \(e_{t_{\tau,i}} = \phi(t_{\tau,i})\) and \(e_{\Delta t_{\tau,i}} = \phi(\Delta t_{\tau,i})\), and sum them to obtain a token-specific conditioning signal \(c_{\tau,i} = e_{t_{\tau,i}} + e_{\Delta t_{\tau,i}} \in \mathbb{R}^d\). Conditioning is applied inside each T-LoopFormer block via an AdaLN-style modulator: a small MLP takes \(c_{\tau,i}\) and outputs \(4d\) parameters per token, which we split into \((\alpha_{\mathrm{msa},\tau,i}, \alpha_{\mathrm{mlp},\tau,i}, \gamma_{\mathrm{msa},\tau,i}, \gamma_{\mathrm{mlp},\tau,i})\). We use RMSNorm (with no learned affinity) before MHSA and FFN, and apply token-wise multiplicative scaling and residual gating as
\[
x \leftarrow x + \alpha_{\mathrm{msa},\tau,i} \odot \mathrm{MHSA}\big( \mathrm{RMSNorm}(x) \odot (1 + \gamma_{\mathrm{msa},\tau,i}) \big),
\]
\[
x \leftarrow x + \alpha_{\mathrm{mlp},\tau,i} \odot \mathrm{FFN}\big( \mathrm{RMSNorm}(x) \odot (1 + \gamma_{\mathrm{mlp},\tau,i}) \big).
\]
The modulator---a SiLU followed by a linear layer of output size \(4d\)---shares weights across tokens, is applied token-wise rather than broadcast over the sequence, and is zero-initialized (weights and bias). This guarantees that the initial behavior equals that of the unmodulated backbone and that conditioning is learned stably.
\section{The difference between gather–scatter during training and inference.}
\label{gather_scatter_diff}
While the gather--scatter scheme is employed in both training and 
inference, its role and mechanics differ between the two phases. These 
differences arise from the distinct objectives of each phase: training 
must maintain a differentiable computation graph, whereas inference must 
maintain a consistent KV cache for decoding.

\paragraph{Purpose of the Scatter.}
\textbf{Training:} The scatter serves as an integral part of the 
differentiable computation graph. After the shared block processes the 
compact tensor $\tilde{H}^{(i)}$, scattering the results back to their 
original positions in $H$ accomplishes three things simultaneously: 
(i) the next loop can gather its own active set $\mathcal{A}_{i+1}$, 
which generally differs from $\mathcal{A}_i$, from a full-length tensor 
whose row indices remain the original sequence positions; (ii) each 
token's state resides at a fixed coordinate indexed by its position, 
enabling correct alignment with the loss; and (iii) gradients flow back 
precisely along the indexed assignment to the corresponding rows of the 
compact tensor. In short, the scatter in training exists primarily to 
keep the computation graph consistent and the gradients well-defined.

\textbf{Inference:} No gradients are involved. The scatter instead 
serves cache management: it updates $H$ with the latest states and, 
crucially, carries the newly computed keys and values to their 
designated slots in the loop-wise KV cache. Writing at the correct 
original positions is essential because future tokens will attend to 
these entries, and the causal alignment of the cache depends on each 
entry residing at its true position.

\paragraph{Address Translation via $\rho^{(i)}$.}
\textbf{Training:} The gather indices $\mathcal{A}_i$ themselves serve 
as the addresses for the scatter, and no additional bookkeeping is 
required: a row at compact index $j$ is written back to position 
$\mathcal{A}_i[j]$.

\textbf{Inference:} Since the KV caches are indexed by original sequence 
positions rather than compact indices, the gather-time mapping 
$\rho^{(i)} = \mathrm{sort}(\mathcal{A}_i)$ must be captured and 
propagated through the shared block to the cache-writing routine. The 
newly computed K/V entries are written into the loop-$i$ cache at their 
original positions $\rho^{(i)}$, not at their compact row indices. This 
positional bookkeeping is an inference-specific burden that training 
does not incur.

\paragraph{Reading the Cache: the Validity Mask $\mu^{(i)}$.}
\textbf{Training:} No cache reads occur, and no masking against stale 
entries is needed --- inactive tokens are simply not present in the 
gathered tensor.

\textbf{Inference:} At loop $i$, attention queries only the loop-$i$ 
cache, whose slots for tokens with $m_\tau < i$ are empty 
(zero-valued). Without masking, the softmax will assign non-negligible 
probability mass to these zero slots. The validity indicator 
$\mu^{(i)}_{t^{'}} = \mathds{1}[\,m_{t^{'}} \geq i\,]$ (plus a block-causal 
mask among currently gathered tokens during batched prefill) is 
therefore mandatory at inference and has no training-time counterpart.

\paragraph{Degeneration during Single-Token Decoding.}
\textbf{Training:} Every forward pass processes the full sequence, so 
gather--scatter always operates on a multi-token tensor and never 
degenerates.

\textbf{Inference:} During autoregressive decoding, only one token is 
processed at a time. Gathering a single row is an identity operation, 
and scattering it back is equivalent to writing in place; the scheme 
therefore degenerates to a pure early exit: the recursion simply 
terminates once the current token leaves the active set. Cache updates 
reduce to ordinary tail concatenation ($\oplus_{t}$), and the address 
translation $\rho^{(i)}$ collapses to the identity.

In essence, the training-time scatter exists to preserve the integrity 
of the computation graph --- coordinates, alignment, and gradients --- 
while the inference-time scatter exists to preserve the integrity of 
the KV cache. The former guarantees that learning is correct; the 
latter guarantees that generation is correct.

\section{Routing Design Considerations}
\label{routing_discussion}
A natural concern regarding the proposed routing mechanism is the :
\emph{non-differentiability} ---
the hard thresholding and discrete index selection involved in gather--scatter
seem to provide no gradient path to the router. We address the concern
below.
\paragraph{Non-Differentiability.}
Although the resulting computation graph is discrete (selected tokens are
gathered via non-differentiable indexing), the router receives gradients
through complementary paths.
\emph{First}, under token-choice routing the selection rule
$m_\tau = \max\{ i : p_\tau^{(i)} \ge \tfrac12 \}$ is derived from the
continue-probability
$\bar{p}_\tau^{(i)} = \sum_{d > i} p_\tau^{(d)}$,
a \emph{differentiable} function of the router logits: gradient signal
flows into the softmax distribution at every loop, even though the binary
mask itself is discrete.
\emph{Second}, for tokens that \emph{are} selected, the task loss
$\mathcal{L}$ backpropagates through the shared block into the
gathered representation and, via the gather indices, reaches the hidden
states on which the router conditions and the router weight matrix itself can
additionally be updated with a straight-through estimator
$\mathrm{STE}[m_\tau] = m_{\tau} + \sigma(s_{\tau}) - \mathrm{sg}(\sigma(s_{\tau}))$ when task-loss
gradients through the routing decision are desired, where $s_{\tau}$ is the "continue probs" at loop-i of token $\tau$. 
\section{Discussion For Performance Degeneration}
Since the number of loops for each token is determined by its initial hidden state (for operational and hardware efficiency—specifically, to enable static tensor compaction via gather–scatter and avoid sequential halting checks at each loop iteration), when the loop count dictated by the token's initial hidden state exceeds the model's maximum loop limit, the decoding results will degenerate. Moreover, because the loop count for each token is not recomputed during the looping stage, the token's understanding of the \textit{context} may be constrained, leading to degraded reasoning performance (a potential limitation of T-LoopFormer).
\label{performance_degeneration}

\end{document}